\documentclass[11pt,a4paper]{article}
\usepackage[hyperref]{acl2019}
\usepackage{times}
\usepackage{latexsym}

\def\@fnsymbol#1{\ensuremath{\ifcase#1\or *\or \dagger\or \ddagger\or
   \mathsection\or \mathparagraph\or \|\or **\or \dagger\dagger
   \or \ddagger\ddagger \else\@ctrerr\fi}}

\aclfinalcopy

\usepackage{microtype}
\usepackage{amsmath}
\usepackage{float}
\usepackage[caption = false]{subfig}
\usepackage[final]{graphicx}
\usepackage[T1]{fontenc} 
\usepackage[utf8]{inputenc}
\usepackage{booktabs} 

\usepackage{hyperref}

\usepackage{url}

\makeatletter
\newcommand{\ssymbol}[1]{^{\@fnsymbol{#1}}}
\makeatother

\title{On Leveraging the Visual Modality for Neural Machine Translation}

\author{Vikas Raunak\textsuperscript{*} Sang Keun Choe\textsuperscript{*} Quanyang Lu\textsuperscript{*} Yi Xu\textsuperscript{*} Florian Metze\\
  Carnegie Mellon University \\
  \texttt{\{vraunak, sangkeuc, qlv, yx2, fmetze\}@andrew.cmu.edu} \\}

\date{}

\begin{document}
\maketitle
\begin{abstract}

Leveraging the visual modality effectively for Neural Machine Translation (NMT) remains an open problem in computational linguistics. Recently, \citeauthor{caglayan2019probing} posit that the observed gains are limited mainly due to the very simple, short, repetitive sentences of the Multi30k dataset (the only multimodal MT dataset available at the time), which renders the source text sufficient for context. In this work, we further investigate this hypothesis on a new large scale multimodal Machine Translation (MMT) dataset, How2, which has 1.57 times longer mean sentence length than Multi30k and no repetition. We propose and evaluate three novel fusion techniques, each of which is designed to ensure the utilization of visual context at different stages of the Sequence-to-Sequence transduction pipeline, even under full linguistic context. However, we still obtain only marginal gains under full linguistic context and posit that visual embeddings extracted from deep vision models (ResNet for Multi30k, ResNext for How2) do not lend themselves to increasing the discriminativeness between the vocabulary elements at token level prediction in NMT. We demonstrate this qualitatively by analyzing attention distribution and quantitatively through Principal Component Analysis, arriving at the conclusion that it is the quality of the visual embeddings rather than the length of sentences, which need to be improved in existing MMT datasets.

\end{abstract}

\section{Introduction}

A number of works have explored integrating the visual modality for Neural Machine Translation (NMT) models, though, there has been relatively modest gains or no gains at all by incorporating the visual modality in the translation pipeline \cite{caglayan2019probing}. In particular, \citet{elliott2017imagination} leverage multi-task learning, \citet{how2} use visual adaptive training, while \citet{mat, mat2, fusion1} use a number of fusion techniques to incorporate features obtained from the visual modality.

Regarding the seemingly low utility of visual modality in machine translation, \citet{bruni2014} hypothesize that the highly relevant visual properties are often not represented by linguistic models because they are too obvious to be explicitly mentioned in text (e.g., birds have wings, violins are brown). Similarly, \citet{louwerse} argue that perceptual information is already sufficiently encoded in textual cues. However, recently \citet{caglayan2019probing} have demonstrated that neural models are capable of leveraging the visual modality for translations, and posit that it is the nature of the Multi30k dataset (the only multimodal machine translation dataset at the time) which is inhibiting gains from the visual modality to emerge, due to the presence of short, simple and repetitive sentences, which renders the source text as sufficient context for translation. In this work, we further investigate this hypothesis on a large-scale multimodal machine translation (MMT) dataset, named How2 \cite{how2}, which has 1.57 times longer sentences, in terms of the mean sentence length, when compared to Multi30k \footnote{Mean Sentence Lengths (in terms of words) are computed post tokenization on the training set for the English language. How2 has a mean sentence length of 20.6, a median of 17, when compared to the mean sentence length of 13 and a median length of 12 for Multi30k \\ \textsuperscript{*} Equal Contribution}.

To this end, we restrict ourselves to the Sequence-to-Sequence (Seq2Seq) framework and propose three simple but novel fusion techniques to ensure the utilization of visual context during different stages (Input Context Encoding, Attention and Supervision) of the Sequence-to-Sequence transduction pipeline. We then evaluate and analyze the results for further insights, with the goal of testing the utility of visual modality for NMT under full source-side linguistic context.

\section{Proposed Fusion Techniques}

In this section, we describe three additions to the Seq2Seq model to ensure that the visual context is utilized at different stages, namely when computing context during each step of the decoder, during attention as well as when computing the supervision signal in the Sequence-to-Sequence pipeline. This is done to encourage the Seq2Seq NMT model to make use of the visual features under full linguistic context. In each case, we assume that the visual features are fine-tuned using a visual encoder, which is trained jointly alongside the Seq2Seq model.

\subsection{Step-Wise Decoder Fusion}

Our first proposed technique is the step-wise decoder fusion of visual features during every prediction step i.e. we concatenate the visual encoding as context at each step of the decoding process. This differs from the usual practice of passing the visual feature only at the beginning of the decoding process \cite{fusion1}. 

\subsection{Multimodal Attention Modulation}


Similar to general attention \cite{luong}, wherein a variable-length alignment vector  $\boldsymbol{a}_{th}(s)$, whose size equals the number of time steps on the source side, is derived by comparing the current target hidden state $h_{t}$ with each source hidden state $\overline{\boldsymbol{h_{s}}}$; we consider a variant wherein the visual encoding $\boldsymbol{v_{t}}$ is used to calculate an attention distribution  $\boldsymbol{a}_{tv}(s)$ over the source encodings as well. Then, the true attention distribution  $\boldsymbol{a}_{t}(s)$ is computed as an interpolation between the visual and text based attention scores. The score function is a content based scoring mechanism as usual.

$$
\boldsymbol{a}_{tv}(s) =\operatorname{align}\left(\boldsymbol{v}_{t}, \overline{\boldsymbol{h}}_{s}\right)
$$
$$
\boldsymbol{a}_{tv}(s) = \cfrac{\exp \left(\operatorname{score}\left(\boldsymbol{v}_{t}, \overline{\boldsymbol{h}}_{s}\right)\right)} {\sum_{s^{\prime}} \exp \left(\operatorname{score}\left(\boldsymbol{v}_{t}, \overline{\boldsymbol{h}}_{s^{\prime}}\right)\right)} 
$$
$$
\operatorname{score}\left(\boldsymbol{v}_{t}, \overline{\boldsymbol{h}}_{s}\right)=\boldsymbol{v}_{t}^{\top} \boldsymbol{W}_{\boldsymbol{v}} \overline{\boldsymbol{h}}_{s}
$$
$$
\boldsymbol{a}_{t}(s) =(1-\gamma) \cdot \boldsymbol{a}_{th}(s) + \gamma \cdot \boldsymbol{a}_{tv}(s)
$$







This formulation differs from \citet{mat} in that we use both the natural language as well as the visual modality to compute attention over the source sentence, rather than having attention over images. Since attention is computed over the same source embeddings (arising from a single encoder) using two different modalities, our approach also differs from \citet{mat2}, which focuses on combining the attention scores of multiple source encoders.

\subsection{Visual-Semantic (VS) Regularizer}

In terms of leveraging the visual modality for supervision, \citet{elliott2017imagination} use multi-task learning to learn grounded representations through image representation prediction. However, to our knowledge, visual-semantic supervision hasn't been much explored for multimodal translation in terms of loss functions.

Our proposed technique is the inclusion of visual-semantic supervision to the machine translation model.
Recently, \citet{ot} proposed an optimal transport based loss function which computes the distance between the word embeddings \footnote{The embeddings are obtained from the Decoder's embedding layer.} of the predicted sentence and the target sentence and uses it as a regularizer $L_{\text{ot}}^{\text{tgt}}$. The purpose of this term is to provide the model with sequence level supervision. We  leverage this idea by including a Cosine distance term, $L_{\text{cosine}}^{\text{visual}}$, between the visual encoding (which is at the sentence level) and the target/predicted sentence embeddings (computed as the average of the target/predicted word embeddings). The purpose of this distance term is to provide sequence level supervision by aligning the visual and text embeddings. In practice, as in \citet{ot}, we introduce a hyperparameter in the loss function:
\begin{displaymath}
L = (1 - \gamma) \cdot L_{\text{mle}} + \gamma \cdot (L_{\text{ot}}^{\text{tgt}} + L_{\text{cosine}}^{\text{visual}}),
\end{displaymath}
where $\gamma$ is a hyper-parameter balancing the effect of loss components (a separate hyperparameter than in Section 2.2).

\section{Results and Analysis}

Throughout our experiments, we use the 300 hours subset of How2 \footnote{ https://github.com/srvk/how2-dataset} dataset \cite{DBLP:journals/corr/abs-1811-00347}, which contains 300 hours of videos, sentence-level time alignments to the ground-truth English subtitles, and Portuguese translations of English subtitles. The How2 dataset has 2048 dimensional pre-trained ResNeXt embeddings \cite {resnext} available for each of the video clips aligned to the sentences.

Further, our baseline model is the canonical Seq2Seq model \cite{seq2seq} consisting of bidirectional LSTM as encoder and decoder, general attention \cite{luong} and length normalization \cite{gnmt}. In all cases, we use the embedding size of 300 and the hidden size of 512. Whenever the visual modality is used, we encode each of the visual features to 300 dimensional vectors through an encoder (consisting of a Linear layer followed by Batch Normalization and ReLU non-linearity) which is also trained end-to-end with the Seq2Seq model. Further, to integrate sequence level supervision as in \citet{ot}, we utilize the Geomloss library \footnote{https://github.com/jeanfeydy/geomloss}, which provides a batched implementation of the Sinkhorn algorithm for the Optimal Transport computation. For all the translation experiments, we preprocess the data by lowercasing and removing the punctuations \cite{how2}, and construct vocabulary at word level. Adam optimizer with a learning rate of 0.001 and a learning rate decay of 0.5 is used to throughout to train our models.

\subsection{Experimental Results}
The performances of the models are summarized in Table \ref{tab:mtresult2}, along with the gains in BLEU points. From Table \ref{tab:mtresult2}, we can make a few observations:
          \begin{table}
          \small
          \begin{tabular}{l c c}
            \toprule
            \textbf{Methods} & \textbf{BLEU} & \textbf{Improvement} \\
            \midrule
            Baseline (En-Pt) & 51.32 &   \\
            + Decoder Fusion (En-Pt) & 51.79 & \textbf{+0.47} \\
            + Multimodal Attention (En-Pt) & 51.85 & \textbf{+0.53}  \\
            + VS Regularization (En-Pt) & 52.00 & \textbf{+0.68}  \\
            \midrule
          \end{tabular}
          \caption{BLEU Score Comparison of the proposed methods}
          \label{tab:mtresult2}
          \vskip -3pt
        \end{table}
        
                  \begin{table}
          \small
          \begin{tabular}{l c c}
            \toprule
            \textbf{Methods} & \textbf{BLEU} & \textbf{Improvement} \\
            \midrule
            Baseline (Pt-En) & 49.12 & \\
            + Decoder Fusion (Pt-En) & 49.68 & \textbf{+0.56} \\
            + Multimodal Attention (Pt-En) & 49.49 & \textbf{+0.37}  \\
            + VS Regularization (Pt-En) & 49.31 & \textbf{+0.19} \\
            \bottomrule
          \end{tabular}
          \caption{BLEU Score Comparison of the proposed methods}
          \label{tab:mtresult4}
          \vskip -3pt
        \end{table}

\begin{enumerate}
    \item The visual modality leads to modest gains in BLEU scores. The proposed VS regularizer leads to slightly higher gain when compared to Decoder-Fusion and Attention modulation techniques for the En-Pt language pair.
    
    \item Further, the gains from incorporating the visual modality are less for Multimodal Attention and VS Regularization in the case of the reversed language pair of Pt-En (Table \ref{tab:mtresult4}), even though the visual modality is common to both the languages. This can possibly be attributed to the How2 dataset creation process wherein first the videos were aligned with English sentences and then the Portuguese translations were created, implying a reduction in correspondence with the visual modality due to errors introduced in the translation process.
\end{enumerate}

\subsection{Discussion}

To analyze the reasons for modest gains, despite incorporating multiple techniques to effectively leverage the visual modality for machine translation, we inspect the dataset as well as the proposed mechanisms.

\begin{figure}[ht!]
    \vskip -3pt
       \includegraphics[width=0.5\textwidth]{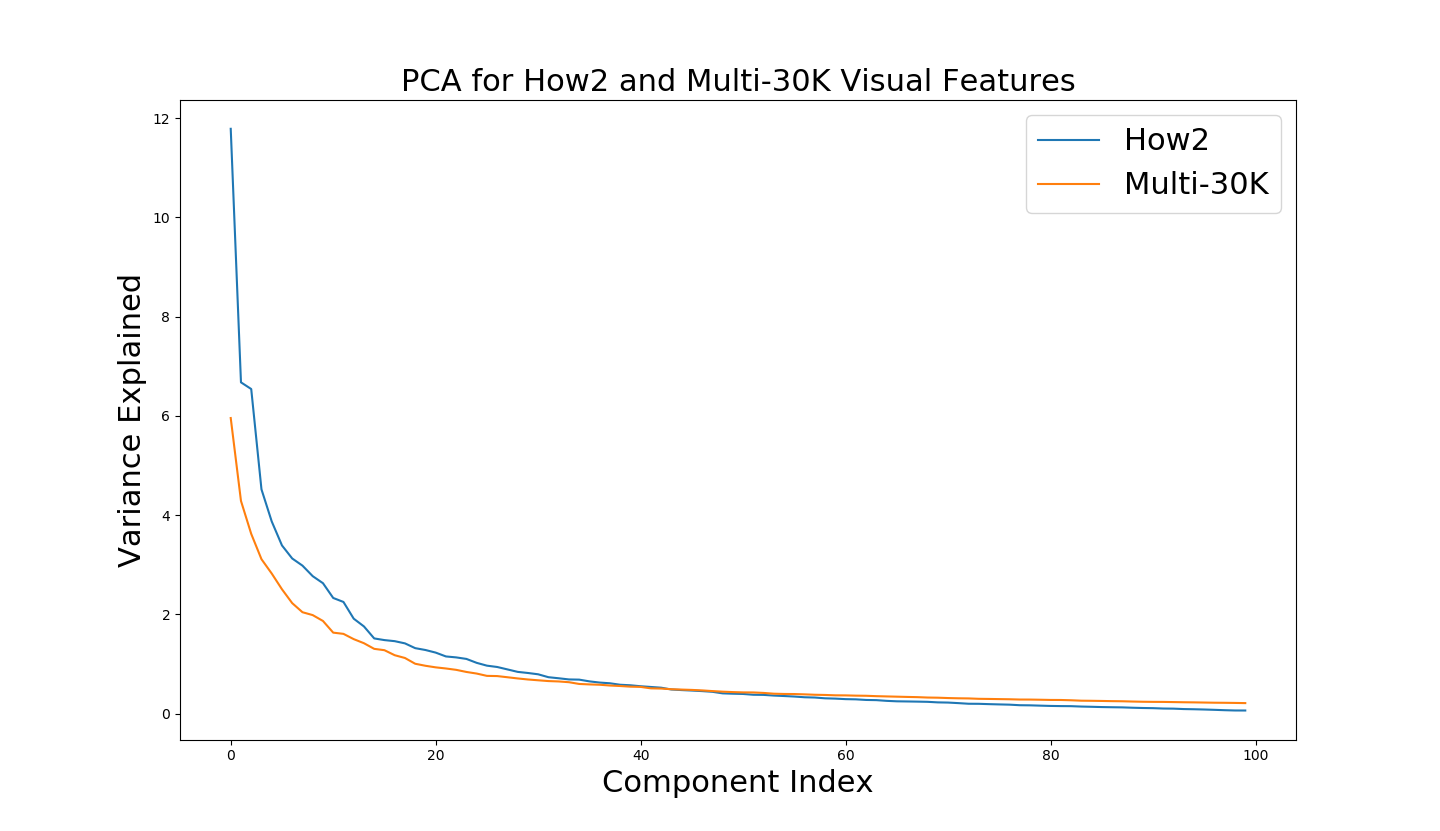}
       
       \includegraphics[width=0.5\textwidth]{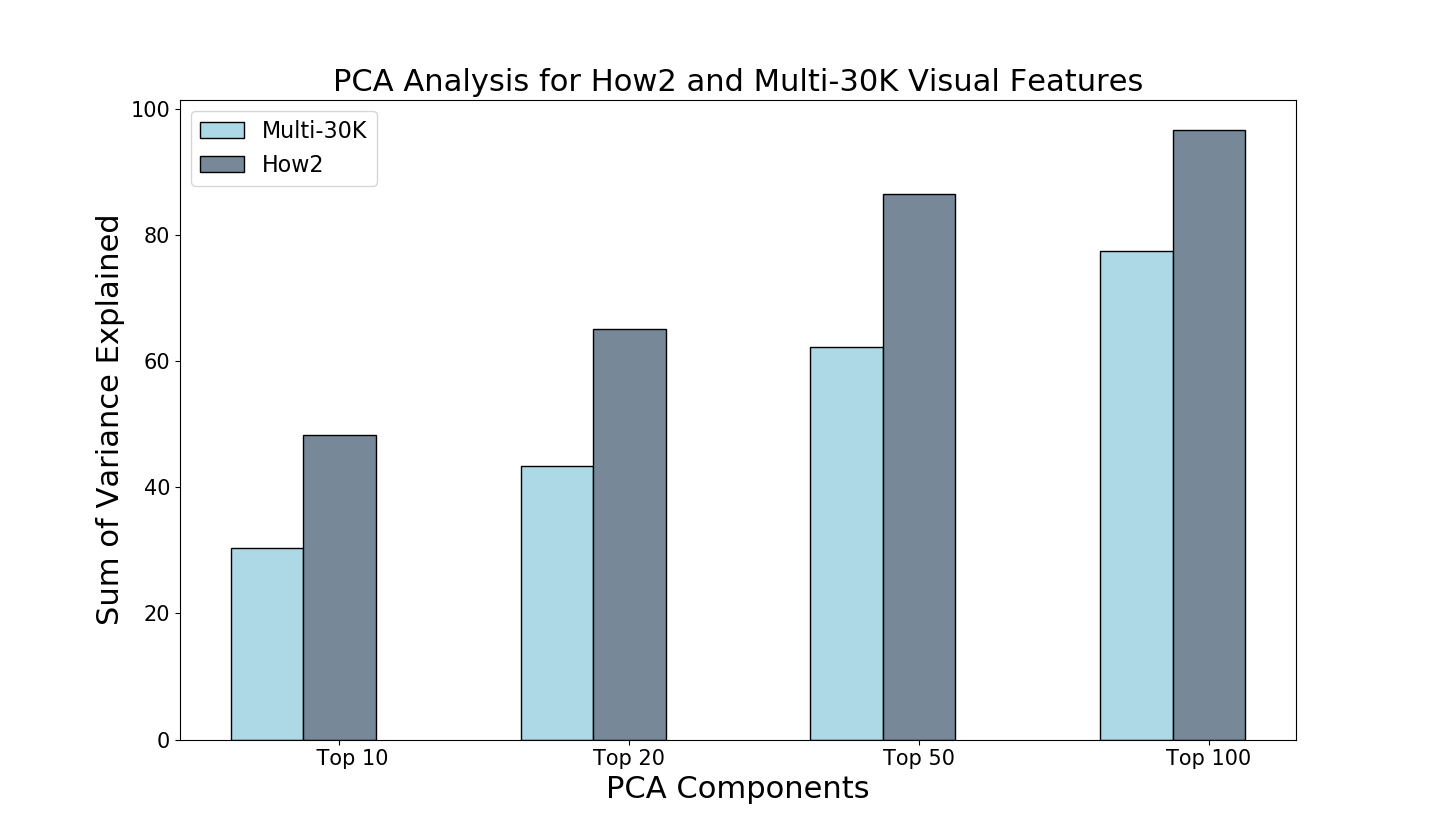}
    \vskip -2pt
    \caption{\textbf{Top}: Variance Explained by the Top 100 Components. \textbf{Bottom}: Cumulative Variance Explained by the Top Components.}
    \label{fig:constraints}
\end{figure}

\subsubsection{PCA of Visual Features}

We first investigate and compare the visual feature quality of the How2 dataset with respect to that of the Multi30k dataset \footnote{https://github.com/multi30k/dataset}. To analyze the discriminativeness of the visual features for both of these datasets, we leverage an analysis mechanism used in \citet{mu} in the context of analyzing word embedding discriminativeness. We analyze the variance of the visual features corresponding to each sentence in the training set. Since the visual features semantically represent the sentence as well, we could analyze how well the features are able to discriminate between the sentences and consequently between the individual words, as a measure of their utility for NMT.

Figure \ref{fig:constraints} (Top) shows the variance explained by the Top 100 principal components, obtained by applying PCA on the How2 and Multi30k training set visual features. The original feature dimensions are 2048 in both the cases. It is clear from the Figure \ref{fig:constraints} that most of the energy of the visual feature space resides in a low-dimensional subspace \cite{mu}. In other words, there exist a few directions in the embedding space which disproportionately explain the variance. These "common" directions affect all of the embeddings in the same way, rendering them less discriminative. Figure \ref{fig:constraints} also shows the cumulative variance explained by Top 10, 20, 50 and 100 principal components respectively. It is clear that the visual features in the case of How2 dataset are much more dominated by the "common" dimensions, when compared to the Multi30k dataset. Further, this analysis is still at the sentence level, i.e. the visual features are much less discriminative among individual sentences, further aggravating the problem at the token level. This suggests that the existing visual features aren't sufficient enough to expect benefits from the visual modality in NMT, since they won't provide discriminativeness among the vocabulary elements at the token level during prediction. Further, this also indicates that under subword vocabulary such as BPE \cite{bpe} or Sentence-Piece \cite{spm}, the utility of such visual embeddings will only aggravate.

\subsubsection{Comparison of Attention Components}

\begin{figure}[ht!]
    \centering
    \vskip -3pt
    \subfloat{\includegraphics[width=0.24\textwidth,height=0.2\textwidth]{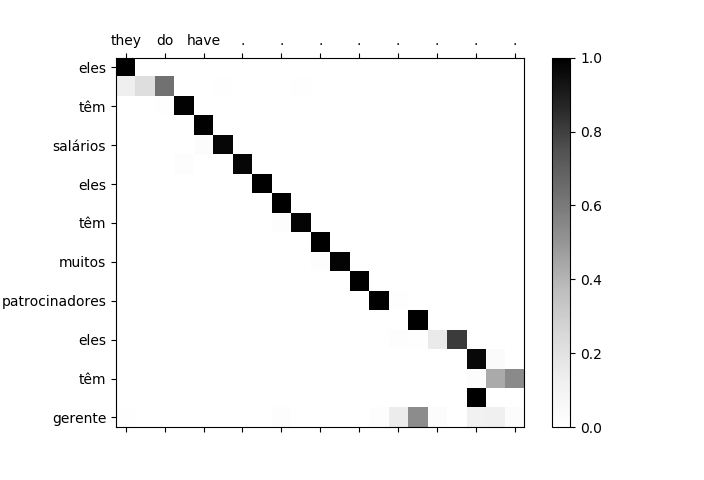}}
    \subfloat{\includegraphics[width=0.24\textwidth,height=0.2\textwidth]{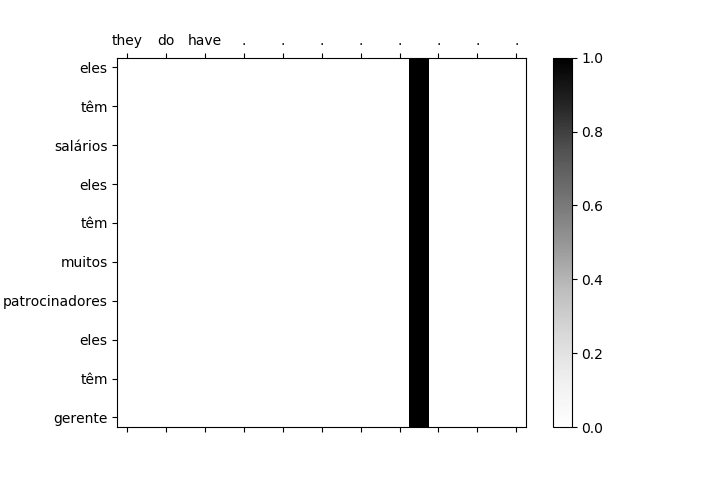}}
    \vskip -7pt
    \caption{\textbf{Left}: Text Based Attention (Horizontal Direction Represents the Source Sentence) \textbf{Right}: Visual Attention for a 21 word Source Sentence (Labels omitted to avoid cluttering).}
    \label{fig:attention1}
\end{figure}

\begin{figure}[ht!]
    \centering
    \vskip -7pt
    \subfloat{\includegraphics[width=0.233\textwidth,height=0.2\textwidth]{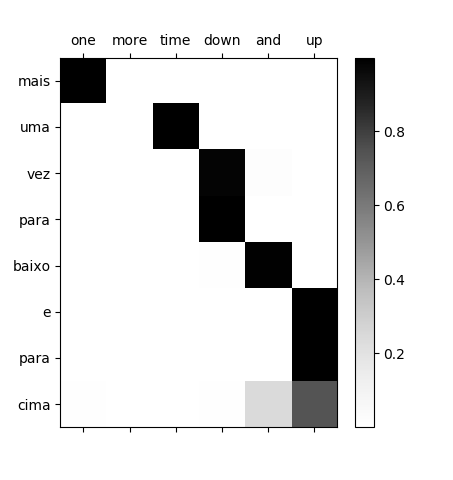}}
    \subfloat{\includegraphics[width=0.233\textwidth,height=0.2\textwidth]{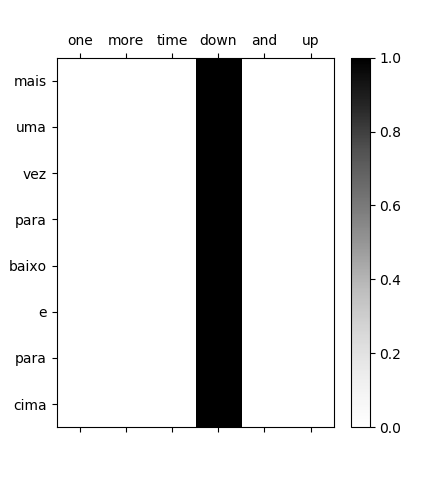}}
    \vskip -7pt
    \caption{\textbf{Left}: Text Based Attention (Horizontal Direction Represents the Source Sentence) \textbf{Right}: Visual Attention for a 7 word Source Sentence.}
    \label{fig:attention2}
\end{figure}

In this section, we analyze the visual and text based attention mechanisms. We find that the visual attention is very sparse, in that just one source encoding is attended to (the maximum visual attention over source encodings, across the test set, has mean 0.99 and standard deviation 0.015), thereby limiting the use of modulation. Thus, in practice, we find that a small weight ($\gamma=0.1$) is necessary to prevent degradation due to this sparse visual attention component. 
Figure \ref{fig:attention1} \& \ref{fig:attention2} shows the comparison of visual and text based attention for two sentences, one long source sentence of length 21 and one short source sentence of length 7.  In both cases, we find that the visual component of the attention hasn't learnt any variation over the source encodings, again suggesting that the visual embeddings do not lend themselves to enhancing token-level discriminativess during prediction. We find this to be consistent across sentences of different lengths.

\section{Conclusions and Future Work}

To conclude, we investigated the utility of visual modality for NMT, under full linguistic context on a new large-scale MMT dataset named How2. Our results on the How2 dataset confirm the general consensus that the visual modality does not lead to any significant gains for NMT, however, unlike \citet{caglayan2019probing} we attribute the relatively modest gains to the limited discriminativeness offered by the existing visual features, rather than the length of the sentences in the dataset. We validate this hypothesis quantitatively through a PCA based analysis of the visual features as well as qualitatively by analyzing attention components. We hope that our work would lead to more useful techniques and better visual features for MMT. An immediate future direction to explore would be to construct more discriminative features for utilizing the visual modality in NMT.

\bibliography{acl2019}
\bibliographystyle{acl_natbib}

\end{document}